\documentclass[a4paper,fleqn]{cas-dc}

\usepackage[numbers,sort&compress]{natbib}

\usepackage{epsfig}
\usepackage{graphicx}
\usepackage{amsmath}
\usepackage{amssymb}
\usepackage{multirow}
\usepackage{color}
\usepackage{amsmath}
\usepackage{graphicx}

\usepackage{booktabs}  

\usepackage{amsmath}
\usepackage{amssymb}
\usepackage{subfigure}
\usepackage{footnote}
\usepackage{epstopdf}
\usepackage{makecell}
\usepackage[ruled]{algorithm2e}
\usepackage{multirow}
\usepackage{soul}
\usepackage{bbding}

\begin{document}
\let\WriteBookmarks\relax
\def\floatpagepagefraction{1}
\def\textpagefraction{.001}

\title[mode = title]{Joint Data and Feature Augmentation for Self-Supervised Representation Learning on Point Clouds}

\tnotemark[1]


\author[1]{Zhuheng Lu}

\address[1]{School of Automation, Nanjing University of Science and Technology, China}

\author[1]{Yuewei~Dai}

\author[2]{	Weiqing~Li}
\address[2]{School of Computer Science and Engineering, Nanjing University of Science and Technology, China}

\author[1]{Zhiyong~Su}


\cortext[cor1]{Corresponding author.
E-mail address: su@njust.edu.cn (Z.Su)}

\begin{abstract}[S U M M A R Y]
 To deal with the exhausting annotations, self-supervised representation learning from unlabeled point clouds has drawn much attention, especially centered on augmentation-based contrastive methods.
   However, specific augmentations hardly produce sufficient transferability to high-level tasks on different datasets. Besides, augmentations on point clouds may also change underlying semantics.
   To address the issues, we propose a simple but efficient augmentation fusion contrastive learning framework to combine data augmentations in Euclidean space and feature augmentations in feature space.
  In particular, we propose a data augmentation method based on sampling and graph generation. 
  Meanwhile, we design a data augmentation network to enable a correspondence of representations by maximizing consistency between augmented graph pairs. We further design a feature augmentation network that encourages the model to learn representations invariant to the perturbations using an encoder perturbation. 
  We comprehensively conduct extensive object classification experiments and object part segmentation experiments to validate the transferability of the proposed framework. 
   Experimental results demonstrate that the proposed framework is effective to learn the point cloud representation in a self-supervised manner, and yields state-of-the-art results in the community. The source code is publicly available at: https://zhiyongsu.github.io/Project/AFSRL.html.

\end{abstract}

\maketitle

\section{Introduction}

Representation learning on point clouds is a challenging task due to the irregular structure and also the need for permutation invariance when processing each point. 
With the rapid development of deep learning-based approaches \cite{qi2017pointnet,qi2017pointnet++,li2018pointcnn}, many works have been proposed for 3D representation problems such as 3D object classification \cite{qi2016volumetric,thomas2019kpconv,wang2019dynamic}, detection \cite{qi2019deep,misra2021end,zhang2020h3dnet} and segmentation \cite{hua2018pointwise,atzmon2018point,tatarchenko2018tangent,wang2019associatively,CAI2019101033}.
Most existing deep neural networks are usually trained in a supervised manner.
However it is often very expensive and time-consuming to collect precise point annotations. 
To remedy this issue, referring to self-supervised learning from unlabeled images \cite{doersch2015unsupervised,chen2021exploring,he2020momentum,liu2019exploiting,ziegler2022self} and videos \cite{sermanet2018time,zhuang2020unsupervised,wang2019self,wang2020self}, self-supervised representation learning from unlabeled point clouds has drawn much attention in the community.

In recent years, numerous self-supervised learning methods have been proposed for point cloud representation learning, which can be divided into generative approaches and discriminative approaches. 
Generative approaches generate labels with the attributes of the data, and are based on reconstruction \cite{sauder2019self,wang2021unsupervised,deng2018ppf,han2019multi,liu2019l2g}, generative models \cite{achlioptas2018learning,wu2016learning}, and other pretext tasks \cite{poursaeed2020self,rao2020global}.
However, generative methods are computationally expensive, and limit the
generality of the learned representations.
Discriminative approaches \cite{du2021self,huang2021spatio,sanghi2020info3d,xie2020pointcontrast,zhang2019unsupervised,zhang2021self} relying on augmentations of point clouds perform discrimination between positive and negative pairs, achieving state-of-the-art results.
Typical point cloud data augmentations in Euclidean space include geometric transformation, point coordinate jittering, subcloud sampling, point dropout, and point-level invariant mapping.
Despite significant progress on contrastive learning approaches based on augmentations, many challenges remain.
First, specific augmentations of point clouds are not suitable for all scenarios because the structural information of the point clouds varies significantly across scenarios, making it difficult to learn the transferable point cloud representations.
Second, even with minimal disruption, it is challenging to maintain semantics adequately throughout typical augmentations.
Therefore, it is highly desirable to seek a transferable and effective learning framework to further improve the performance of point cloud representations.

This paper proposes an Augmentation Fusion Self-Supervised Representation Learning (AFSRL) framework by combining data augmentations in Euclidean space and feature augmentations in feature space to construct a stable and invariant point cloud representation.
The AFSRL is composed of three modules: data augmentation module, data augmentation network, and feature augmentation network.
The data augmentation module first generates augmented graph pairs as the input of the data augmentation network.
The goal of the data augmentation network is to capture the correspondence between the augmented pairs. 
The feature augmentation network is inspired by the observation \cite{xia2022simgrace} that graph data
can maintain their semantics effectively during encoder perturbations without necessitating manual trial-and-error, time-consuming search, or expensive domain expertise.
Specifically, we take one generated graph as input and a perturbed GNN model as the encoder to obtain the augmented feature representation. 
Moreover, we consider maximizing the consistency of the corresponding representations. 
With the encoder perturbation as noise, we can obtain two different feature embeddings for the same input as positive pairs. 
The perturbation cues boost the performance of learned representations.

To evaluate the proposed self-supervised representation learning framework, we adopt the learned representation for object classification and part segmentation tasks. 
We confirm that the learned representations are easily transferable to downstream tasks directly by pre-training on large datasets.
According to the experimental results, we attain new state-of-the-art performances among all these tasks.

In summary, our main contributions in this paper are as follows:

\begin{itemize}
	
	\item  We introduce an augmentation fusion framework that imposes invariance to data augmentation, and simultaneously feature augmentation encourages the model to learn representations invariant to the perturbations. 
	
	\item We demonstrate that the transferability of learned representations, which can be readily adapted to downstream tasks on different datasets.
	
	\item Experimental results show that compared with other unsupervised methods, our AFSRL achieves competitive results and narrows the gap between unsupervised methods and supervised methods on a series of datasets.

\end{itemize} 	

The rest of the paper is organized as follows. Section \ref{sec:related} briefly reviews the related work on supervised representation learning on point clouds and self-supervised learning on point clouds. In Section \ref{sec:method}, an overview of the framework is provided. Then, the details of the architecture are introduced. After that, experimental results are presented in Section \ref{sec:experiment}. Finally, conclusions and recommendations for future research are given in Section \ref{sec:Conclusion}.

\begin{figure*}[!t]
	\centering
	\includegraphics[width=0.9\textwidth]{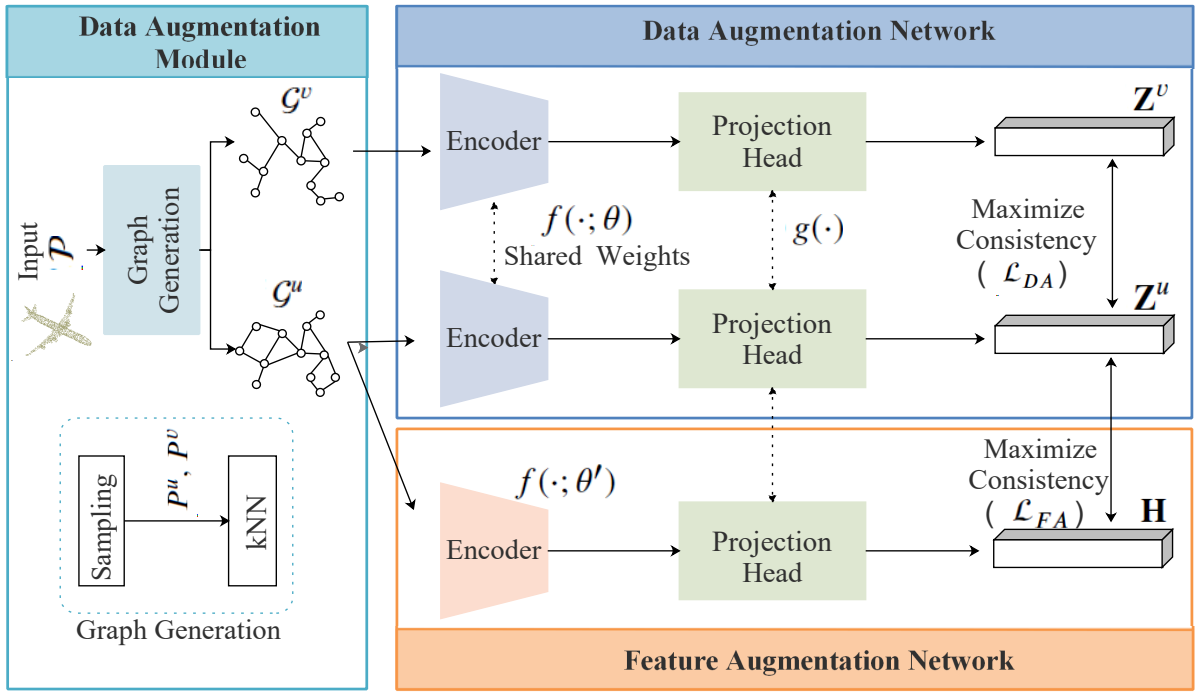}\\
	\caption{The architecture of the proposed AFSRL for self-supervised point cloud representation learning. The input 3D point clouds are firstly generated into graph pairs. Data augmentation network inputs augmented graph pairs to a shared GNN encoder and maximizes the similarity of the representation pairs. Moreover, feature augmentation network takes	one generated graph as input and the perturbed GNN as encoder to obtain the correlated representation. The AFSRL trains the model by combining the learning objectives of the networks.}
	\label{fig:framework}
\end{figure*}

\section{Related Work}
\label{sec:related}
In this section, we review a number of previous works on supervised representation learning and self-supervised learning on point clouds.

\subsection{Supervised Representation Learning on Point Clouds}
Many approaches have been proposed to address various tasks on unordered point cloud representations, such as object classification \cite{qi2016volumetric,li2018pointcnn,liu2019relation,qi2017pointnet++,thomas2019kpconv,wang2019dynamic,CHEKIR2022101164}, object detection \cite{qi2019deep,misra2021end,zhang2020h3dnet,pham2019jsis3d,shi2019pointrcnn,yi2019gspn}, and segmentation \cite{hua2018pointwise,atzmon2018point,xie2018attentional,xu2018spidercnn,klokov2017escape,tatarchenko2018tangent,wang2019associatively,ZHANG2022101120}.
Approaches on point cloud supervised representation learning can
be mainly classified into three categories: point-based, view-based, and voxel-based.

Among the point-based approaches, one pioneer method PointNet \cite{qi2017pointnet} was designed to directly feed raw point clouds into neural networks, and obtain the global point cloud feature. The framework can be used for point cloud segmentation, part segmentation, semantic classification, and other tasks. 
Since then, numerous advancements have been made in point-based tasks. 
Qi et al. \cite{qi2017pointnet++} proposed PointNet++ to hierarchically aggregate the local features of the point cloud.
Li et al. \cite{li2018pointcnn} proposed PointCNN which uses $\mathcal{X}$-Conv operator on the input point cloud, and then applies a typical convolution operator on the transformed features.
Another branch of point-based approaches relies on graph convolutions on 3D point clouds.
Wang et al. \cite{wang2019dynamic} proposed DGCNN using an edge-convolution network
(EdgeConv) to specifically model local neighborhood information.
Landrieu et al. \cite{landrieu2018large} generate a superpoint graph of a point cloud and
learn the 3D point geometric organization.

View-based methods generate a shape descriptor from multiple views of each point cloud.
Su et al. \cite{su2015multi} proposed MVCNN to apply multi-view convolutional neural network for 3D recognition. The 3D shapes were rendered in different views, each of which is passed through a unified CNN architecture. 
Xie et al. \cite{xie2015projective} generated 2.5D depth images and proposed an extreme learning machine to achieve projective feature learning for 3D shapes.
He et al. \cite{he2020improved} proposed a combination scheme of group-view similarity learning and adaptive margin-based triplet-center loss to improve MVCNN for 3D shape retrieval

Voxel-based methods use convolutional neural networks to extract 3D shape descriptors from regular and dense 3D shape voxel. 
Wu et al. \cite{wu20153d} proposed 3D ShapeNets to learn the  hierarchical compositional part representation of complex 3D shapes from 3D shape volumetric data.
Maturana et al. \cite{maturana2015voxnet} proposed VoxNet to integrate a volumetric representation with a supervised 3D Convolutional Neural Network.
Le et al. \cite{le2018pointgrid} proposed PointGrid which incorporates a constant number of points in each grid cell thus allowing the network to better capture local geometric details.

\subsection{Self-supervised Learning on Point Clouds}

Recently, many approaches have sought to explore semantic information in an unsupervised manner with designed pretext tasks. Existing self-supervised learning on point clouds can be classified as either generative or discriminative. 

Generative approaches build a distribution over data and latent embedding and use the learned embeddings as point cloud representations. These generative models use generative adversarial networks \cite{wu2016learning,achlioptas2018learning,han2019view} or auto-encoders \cite{li2018so,yang2018foldingnet,han2019multi,eckart2021self,zhao20193d}. However, generative approaches are computationally expensive, and the learning of generalizable representation unnecessarily
relies on recovering such high-level details.

Among discriminative methods, contrastive methods \cite{du2021self,huang2021spatio,sanghi2020info3d,xie2020pointcontrast,zhang2019unsupervised,zhang2021self} currently achieve state-of-the-art performance in self-supervised learning. Different from generative approaches that maximize the data likelihood, the contrastive learning-based methods typically learn representation by maximizing feature similarity between differently augmented pairs.
Typical point cloud data augmentations include geometric transformation; point coordinate jittering; subcloud sampling; point dropout; and point-level invariant mapping.
Zhang et al. \cite{zhang2019unsupervised} proposed the part contrast by segmenting one object into two parts to learn features from unlabeled point clouds.
Afham et al. \cite{afham2022crosspoint} proposed a cross-modal contrastive learning approach to encourage the 2D image feature to be embedded close to the corresponding 3D point cloud.
Xie et al. \cite{xie2020pointcontrast} proposed PointContrast on two transformed views of the given point cloud with a contrastive loss.
Despite the success, choosing an augmentation still requires the time-consuming manual trial-and-error, laborious search, or expensive domain knowledge. 
Instead, our AFSRL breaks through typical data augmentation methods as prerequisite and narrow the augmentation gap mentioned above.

\section{Method}
\label{sec:method}

\subsection{Overview}
Figure \ref{fig:framework} illustrates our AFSRL architecture, which consists data augmentation module, data augmentation network, and feature augmentation network.
The augmentation module first generates graphs via sampling and $K$-nearest neighbor as the data augmentation.
The data augmentation network inputs augmented graph pairs $\mathcal{G}^{u}$ and $\mathcal{G}^{v}$ to a shared GNN encoder and maximizes the similarity of the representation pairs. 
Meanwhile, the feature augmentation network takes one graph from the augmented graph pairs as input, and employs the perturbed GNN as the encoder to obtain the augmented feature representation. 
The feature augmentation network maximizes the consistency of correlated representations using contrastive learning loss functions. 
The AFSRL trains the model by combining the learning objectives of two networks.

\subsection{Data Augmentation Module}
\label{sec.augmentation}
The data augmentation module obtains positive pairs of the point clouds as the input of the following networks in an unsupervised setting.
This can be thought of as a form of data augmentation, which is an essential component of our framework.
First, given a point cloud $\mathcal{P}$, we randomly sample two versions of point cloud $P^{u}$, $P^{v}$ with a uniform scale.
Then, we generate the $K$-nearest neighbor adjacency graphs $\mathcal{G}^{u}$, $\mathcal{G}^{v}$ of the sampled point clouds as the augmented pairs of the input point cloud $\mathcal{P}$.
Our method differs from traditional data augmentations by explicitly preserving the semantic structure to create better embeddings.

\subsection{Data Augmentation Network}
The data augmentation network consists of a shared encoder and a shared projection head. 
Inspired by recent contrastive learning algorithms, the data augmentation network learns representations
by maximizing consistency between differently augmented
graphs of the same point cloud via a contrastive loss in
the latent space.

First, following the architecture in \cite{grill2020bootstrap,xia2022simgrace}, a backbone GNN encoder $f(\cdot;\theta)$ is used to extract representation vectors, which are denoted as $\mathbf{Y}^{u}$ and $\mathbf{Y}^{v}$, from augmented graph examples,
 
\begin{equation}
\mathbf{Y}^{u}=f(\mathcal{G}^{u};\theta),\ \mathbf{Y}^{v}=f(\mathcal{G}^{v};\theta).
\end{equation}
Any deep GNN can be used as the feature extractor, such as GCN \cite{kipf2016semi}, GAT \cite{velivckovic2017graph}, and GRAPHSAGE \cite{hamilton2017inductive}.

Then, a neural network projection head $g(\cdot)$ is provided to map representations to the space where contrastive loss is applied. 
We denote the projected vectors of $\mathbf{Y}^{u}$ and $\mathbf{Y}^{v}$ as $\mathbf{Z}^{u}$ and $\mathbf{Z}^{v}$, respectively. 
As proved by several previous approaches \cite{chen2020simple,xia2022simgrace}, a projection head can enhance performance.
By leveraging
the nonlinear transformation $g(\cdot)$, more information can
be formed and maintained in the representation to prevent information loss
induced by the contrastive loss.
Besides, a nonlinear projection is better than a linear projection.
Therefore, in this framework, we adopt a multi-layer perceptron (MLP) to obtain the projected vectors $\mathbf{Z}^{u}$ and $\mathbf{Z}^{v}$,
\begin{equation}
\mathbf{Z}^{u}=g(\mathbf{Y}^{u}),\mathbf{Z}^{v}=g(\mathbf{Y}^{v}).
\end{equation}

The goal is to maximize the similarity of $\mathbf{Z}^{u}$ with $\mathbf{Z}^{v}$ while minimizing the similarity with all the other projected vectors in the minibatch of graphs.
The AFSRL is then trained by minimizing normalized temperature-scaled cross entropy loss (NT-Xent) \cite{wu2018unsupervised,xia2022simgrace,you2020graph} to maximize the consistency of $\mathbf{Z}^{u}$ and $\mathbf{Z}^{v}$.
We compute the cosine similarity between $\mathbf{Z}^{u}$ and $\mathbf{Z}^{v}$ as follows:

\begin{equation}
s(\mathbf{Z}^{u},\mathbf{Z}^{v})=\frac{{\mathbf{Z}^{u}}^{\intercal}\mathbf{Z}^{v}}{\parallel \mathbf{Z}^{u} \parallel \ \parallel \mathbf{Z}^{v} \parallel}.
\end{equation}

We compute the loss function $l$ for the $n$-th augmented pair of examples as:

\begin{footnotesize}
\begin{equation}
l(n,\mathbf{Z}^{u},\mathbf{Z}^{v})=-\log\frac{\exp(s(\mathbf{Z}_{n}^{u},\mathbf{Z}_{n}^{v})/\tau)}{\sum\limits_{\substack{k=1\\k\ne n}}^{N}\exp(s(\mathbf{Z}_{n}^{u},\mathbf{Z}_{k}^{u})/\tau)+\sum\limits^{N}_{k=1}\exp(s(\mathbf{Z}_{n}^{u},\mathbf{Z}_{k}^{v})/\tau)},
\end{equation}
\end{footnotesize}where $\tau$ is the temperature parameter, $N$ is the minibatch size. The final loss $\mathcal{L}_{DA}$ is computed across all positive pairs in the minibatch:

\begin{equation}
\mathcal{L}_{DA}=\frac{1}{2N}\sum^{N}_{n=1}[l(n,\mathbf{Z}^{u},\mathbf{Z}^{v})+l(n,\mathbf{Z}^{v},\mathbf{Z}^{u})].
\end{equation}

\subsection{Feature Augmentation Network}

We introduce feature augmentation network to learn the representations through the interactions of two branches, thus yielding better representation learning capability of point clouds. The feature augmentation network has a similar architecture as the data augmentation network, but uses a different set of weights.

More precisely, we first obtain a corresponding perturbed encoder $f(\cdot;\theta')$ of the backbone GNN encoder $f(\cdot;\theta)$. The perturbed encoder $f(\cdot;\theta')$ computes the representations $\mathbf{R}$ for each input graph $\mathcal{G}$, which can be described as 
\begin{equation}
\mathbf{R}=f(\mathcal{G};\theta').
\end{equation}

The feature augmentation network perturbs the encoder with a random Gaussian noise instead of momentum updating \cite{grill2020bootstrap}. 
Therefore, the perturbed encoder $f(\theta')$ can be mathematically described as 
\begin{equation}
\theta_{l}^{'}=\theta_{l}+\epsilon\cdot\hat{\theta_{l}},
\end{equation}where $\theta_{l}$ is the weight tensor of the $l$-th layer of the GNN encoder, $\theta_{l}^{'}$ is the weight tensor of the corresponding perturbed version, $\epsilon$ is a parameter that controls the scale of the perturbation, and $\hat{\theta_{l}} $ denotes the perturbation
term that samples from the Gaussian distribution $\mathcal{N}(0,\sigma_{l}^{2})$.

Afterwards, the representation $\mathbf{R}$ is projected to a latent space by a non-linear transformation $g(\cdot)$, 
\begin{equation}
\mathbf{H}=g(\mathbf{R}).
\end{equation}

The AFSRL is then trained by minimizing normalized temperature-scaled cross entropy loss to maximize the consistency of $\mathbf{Z}$ and $\mathbf{H}$.
Then the loss function for a perturbed pair of examples is defined as
\begin{footnotesize}
	\begin{equation}
	l(n,\mathbf{Z},\mathbf{H})=-\log\frac{\exp(s(\mathbf{Z}_{n},\mathbf{H}_{n})/\tau)}{\sum\limits_{\substack{k=1\\k\ne n}}^{N}\exp(s(\mathbf{Z}_{n},\mathbf{Z}_{k})/\tau)+\sum\limits^{N}_{k=1}\exp(s(\mathbf{Z}_{n},\mathbf{H}_{k})/\tau)},
	\end{equation}
\end{footnotesize}
where $\tau$ is the temperature parameter. The final loss function $\mathcal{L}_{FA}$ is computed across all positive pairs in the minibatch, 
\begin{equation}
\mathcal{L}_{FA}=\frac{1}{2N}\sum^{N}_{n=1}[l(n,\mathbf{Z},\mathbf{H})+l(n,\mathbf{H},\mathbf{Z})].
\end{equation}

\begin{algorithm}[htp]
	\caption{Main learning algorithm of AFSRL}
	\label {alg:AFSRL}	
	\LinesNumbered 
	\KwIn{Point cloud set $\mathcal{D}=\{ \mathcal{P}_{1},\mathcal{P}_{2},\cdots,\mathcal{P}_{L} \} $, batch size $N$, initial encoder weights $ \theta $\\}

	\For{ each point cloud}{
		Sample two augmentation $\mathcal{G}^{u} $, $\mathcal{G}^{v} $
		\textcolor{blue}	{// \texttt{data augmentation\\}}
		\For{ each mini-batch}
		{$\mathbf{Y}^{u}=f(\mathcal{G}^{u};\theta), \mathbf{Y}^{v}=f(\mathcal{G}^{v};\theta)$	
			\textcolor{blue}	{// \texttt{encode\\}}	
			$\mathbf{Z}^{u}=g(y^{u}), \mathbf{Z}=g(y^{v})$
			\textcolor{blue}	{// \texttt{project\\}}	
			$	\mathcal{L}_{DA}=\frac{1}{2N}\sum^{N}_{n=1}[l(n,\mathbf{Z}^{u},\mathbf{Z}^{v})+l(n,\mathbf{Z}^{v},\mathbf{Z}^{u})]	$	\textcolor{blue}	{// \texttt{compute loss\\}}
				$\mathbf{R}=f(\mathcal{G};\theta')$	
				\textcolor{blue}	{// \texttt{encode\\}}	
				$\mathbf{H}=g(\mathbf{R})$
				\textcolor{blue}	{// \texttt{project\\}}	
				$
\mathcal{L}_{FA}=\frac{1}{2N}\sum^{N}_{n=1}[l(n,\mathbf{Z},\mathbf{H})+l(n,\mathbf{H},\mathbf{Z})]
				$	\textcolor{blue}	{// \texttt{compute loss\\}}
				$\mathcal{L}=\alpha\mathcal{L}_{DA}+\beta\mathcal{L}_{FA}$
		}
		Update weights
	}
	
\end{algorithm}

\subsection{Overall Objective}

Finally, we obtain the loss function during training with the combination of $\mathcal{L}_{DA}$ and $\mathcal{L}_{FA}$, where $\mathcal{L}_{DA}$ imposes invariance to data augmentation
while $\mathcal{L}_{FA}$ enforces the model to learn representations
invariant to the perturbations,

\begin{equation}
\mathcal{L}=\alpha\mathcal{L}_{DA}+\beta\mathcal{L}_{FA},
\end{equation}
where the constant $\alpha$ and $\beta$ balance the contrastive loss $\mathcal{L}_{DA}$ and $\mathcal{L}_{FA}$, respectively. Algorithm \ref{alg:AFSRL} details the proposed AFSRL.

\section{Experimental Results}
\label{sec:experiment}
In this section, we evaluate the feature representations of point clouds learned with the proposed AFSRL for object classification and part segmentation task. We first introduce the datasets and implementation details, and then present the experimental results.

\subsection{Pre-training}

\textbf{Dataset.} We use  the ShapeNet \cite{chang2015shapenet} for learning the self-supervised representation, which consists of 57448 objects from 55 categories. 
By augmenting each point cloud into two relevant graphs defined in Section \ref{sec.augmentation}, we generate positive pairs of point clouds as the input.

\textbf{Implementation Details.} We employ the GCN proposed in \cite{kipf2016semi} as the feature extractor. The encoder architecture is defined as:
\begin{equation}\label{eq.propagation}
\mathbf{X}^{(l)}=\sigma(\mathbf{D}^{-1/2}\mathbf{A}\mathbf{D}^{-1/2}\mathbf{X}\mathbf{W}^{(l)}),
\end{equation}
where $\mathbf{X}^{(l)}$ is the node embedding matrix of the $l$-th layer, $\mathbf{D}$ is the degree matrix, $\mathbf{A}$ is the adjacency matrix, $\sigma(\cdot)$ is a nonlinear activation function such as ReLU, and $\mathbf{W}$ is the trainable weight matrix for the $l$-th layer. 
We employ a 2-layer MLP as the projection heads. 
$\alpha$ and $\beta$ are set as equal weight in the training phase.

\begin{table}[]
	\renewcommand\arraystretch{1.5}
	\caption{Comparison of classification results with existing methods on ModelNet40.}
	\centering
	\begin{tabular}{lc}
		\hline
		Method & Accuracy (\%)\\ \hline
		3D-GAN \cite{wu2016learning}    &      83.3    \\
		Latent-GAN \cite{achlioptas2018learning} &    85.7      \\
		SO-Net \cite{li2018so}  &      87.3    \\
		MRTNet \cite{han2019multi}  &    86.4      \\
		3D-PointCapsNet \cite{zhao20193d} &    88.9      \\
		FoldingNet \cite{yang2018foldingnet}  &    88.4      \\
		ClusterNet \cite{zhang2019unsupervised}  &   86.8       \\
		DepthContrast \cite{zhang2021self}  &    85.4      \\
		Sauder et al. + PointNet \cite{sauder2019self}  &     87.3     \\ 
		Sauder et al. + DGCNN \cite{sauder2019self}   &     89.1     \\
		Poursaeed et al. + PointNet \cite{poursaeed2020self}  &    88.6      \\
		Poursaeed et al. + DGCNN \cite{poursaeed2020self}   &      90.7    \\
		CrossPoint + PointNet \cite{afham2022crosspoint}  &     89.1     \\ 
		CrossPoint + DGCNN \cite{afham2022crosspoint}   &     91.2     \\
		STRL + PointNet \cite{huang2021spatio}  &      88.3    \\ 
		STRL + DGCNN \cite{huang2021spatio}   &      90.9    \\
		AFSRL( Ours )  &    \textbf{91.5}      \\
		\hline
	\end{tabular}
	\label{tab:ModelNet40}
\end{table}

%

\subsection{Object Classification}
We evaluate our classification experiments on ModelNet40 \cite{wu20153d} and ScanObjectNN \cite{uy2019revisiting}. 
ModelNet40 is a synthetic dataset obtained by sampling 3D CAD models. 
It includes 12331 objects from 40 categories, and the dataset is split into 9843 examples for training and 2468 for testing.
ScanObjectNN is a real scanned 3D point cloud classification dataset. 
It contains 2880 objects from 15 categories, 2304 for training and 576 for testing.

We follow the same protocols in \cite{huang2021spatio,afham2022crosspoint} to train a linear Support Vector Machine (SVM) using the encoded feature representations. We randomly sample 1024 points from each object for both training and testing the classification results.

Table \ref{tab:ModelNet40} tabulates the linear classification results on ModelNet40 benchmarks. 
Our method achieves an object classification accuracy of 91.4\%. 
We compare our method with the results of state-of-the-art methods in terms of class average accuracy. 
The proposed method outperforms previous state-of-the-art unsupervised methods on the ModelNet40 dataset, which justifies the effectiveness of our method.

\begin{table}[]
	\renewcommand\arraystretch{1.5}
	\caption{Comparison of classification results with existing methods on ScanObjectNN.}
	\centering
	\begin{tabular}{lc}
		\hline
		Method & Accuracy (\%)\\ \hline
		
		Sauder et al. + PointNet \cite{sauder2019self}  &     55.2     \\ 
		Sauder et al. + DGCNN \cite{sauder2019self}   &     59.5   \\
		OcCo + PointNet \cite{wang2021unsupervised}  &    69.5     \\
		OcCo + DGCNN \cite{wang2021unsupervised}   &      78.3    \\
		CrossPoint + PointNet \cite{afham2022crosspoint}  &     75.6     \\ 
		CrossPoint + DGCNN \cite{afham2022crosspoint}   &     81.7    \\
		STRL + PointNet \cite{huang2021spatio}  &      74.2    \\ 
		STRL + DGCNN \cite{huang2021spatio}   &      77.9    \\
		AFSRL( Ours )  &    \textbf{82.3}      \\
		\hline
	\end{tabular}
	\label{tab:ScanObjectNN}
\end{table}

Table \ref{tab:ScanObjectNN} reports the linear classification results on ScanObjectNN benchmarks. 
Our method achieves an object classification accuracy of 82.1\%.
The proposed AFSRL also outperforms all the state-of-the-art unsupervised and self-supervised methods on ScanObjectNN. 
Our method is extremely effective in extracting discriminant features from the results.

\subsection{Object Part Segmentation}
We transfer the pre-trained model to the object part segmentation experiments to further validate that our approach is qualified for point cloud representation learning. 
The goal of object part segmentation is to predict the semantic part label for each point of input point
clouds.
For this task, the ShapeNetPart dataset \cite{yi2016scalable} is used as the benchmark dataset.
The ShapeNetPart benchmark dataset consists of 16881 3D objects from 16 categories. 
We employ 12137 models for training, and 2874 models for testing following PointNet++.
We use the mean Intersection-over-Union (IoU) as the evaluation metric.
IoU is computed between ground truth and the prediction for each part of an object. 
The mean IoU (mIoU) is then calculated by averaging the IoUs of all test objects.

\begin{figure*}[!t]
	\centering
	\includegraphics[width=\textwidth]{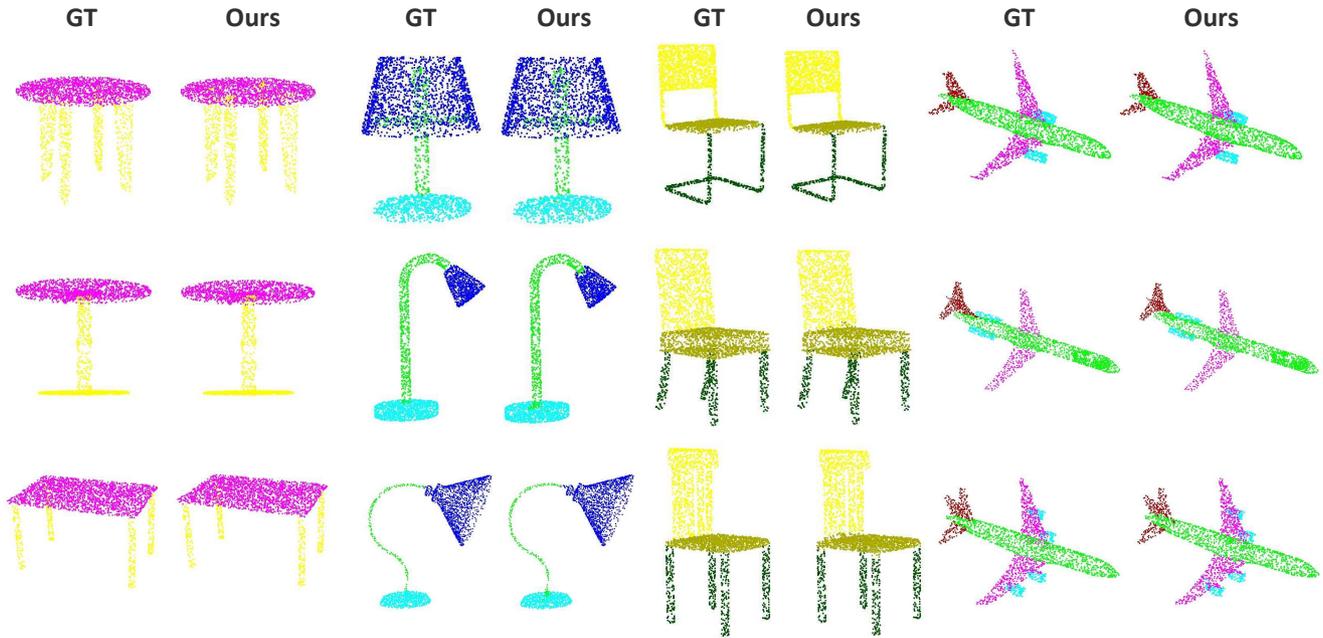}\\
	\caption{Qualitative results of part segmentation on the ShapeNetPart dataset. In each object pair, the left column is the ground truth (GT), and right column is our segmentation result, where parts with the same color have a consistent meaning. Four categories of objects are table, lamp, chair, and airplane.} 
	\label{fig:resultShapeNetPart}
\end{figure*}

We compare our approach with both unsupervised approaches and supervised approaches, as tabulated in Table \ref{tab:ShapeNetPart}.
Compared with the stated-of-the-art unsupervised methods, the AFSRL acquires the best mean IoU of 85.6\%.
This is mainly due to the fact that the proposed augmentation fusion method preserves semantic information better.
Figure \ref{fig:resultShapeNetPart} visualizes some examples of our segmentation results, where the segmentation results are highly consistent with the ground truth.

\begin{table}[]
	\renewcommand\arraystretch{1.5}
	\caption{Comparison of part segmentation results with existing methods on ShapeNetPart dataset.}
	\centering
	\begin{tabular}{l|c|c}
		\hline
		Method & \multicolumn{1}{l|}{Supervision}                  &  Mean IoU (\%)\\ \hline
		PointNet \cite{qi2017pointnet}   & \multirow{9}{*}{Supervised}            &     83.7     \\
		PointNet++ \cite{qi2017pointnet++}     &                                        &    85.1      \\
		PointCNN \cite{li2018pointcnn}   &                                        &   86.1       \\
		DGCNN \cite{wang2019dynamic}    &                                        &      85.1    \\
		KD-Net \cite{klokov2017escape}   &                                        &    82.3      \\
		Point2Sequence \cite{liu2019point2sequence}   &                                        &      85.2    \\ 
		DensePoint \cite{liu2019densepoint}   &                                        &      86.4    \\ 
		PointTransformer \cite{zhao2021point}   &                                        &    \textbf{86.6}      \\
		Stratified Transformer \cite{lai2022stratified}   &                                        &    \textbf{86.6}      \\ \hline
		Self-Contrast \cite{du2021self}   & \multicolumn{1}{l|}{\multirow{7}{*}{Self-Supervised}} &    82.3      \\
		Sauder et al. \cite{sauder2019self}   & \multicolumn{1}{l|}{}                  &  85.3        \\
		OcCo \cite{wang2021unsupervised}   & \multicolumn{1}{l|}{}                  &     85.0     \\
		PointContrast \cite{xie2020pointcontrast}   & \multicolumn{1}{l|}{}                  &     85.1     \\
		Liu et al. \cite{liu2021point}   & \multicolumn{1}{l|}{}                  &     85.3     \\
		CrossPoint \cite{afham2022crosspoint}   & \multicolumn{1}{l|}{}                  &   85.5       \\
		AFSRL    & \multicolumn{1}{l|}{}                  &    \textbf{85.7}     \\
		\hline
	\end{tabular}
	\label{tab:ShapeNetPart}
\end{table}

\subsection{Ablations and Analysis}
\begin{figure}[!t]
	\centering
	\includegraphics[width=0.45\textwidth]{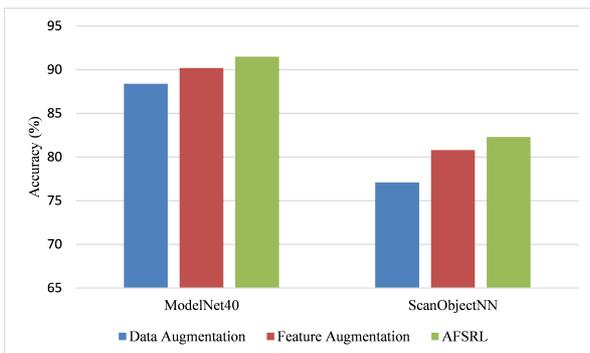}\\
	\caption{Effect of each network of the proposed framework for the task of object classification on ModelNet40 and ScanObjectNN datasets in terms of accuracy.} 
	\label{fig:model}
\end{figure}

\textbf{Impact of the fusion learning objective.} We first evaluate the importance of the two networks proposed in this framework. 
Figure \ref{fig:model} shows the impact of the data augmentation network and feature augmentation network for the task of object classification on ModelNet40 and ScanObjectNN datasets.
We find that the combination of data augmentation network and feature augmentation network contribute to better learned representations than they do separately.
The data augmentation network network enforces the model to capture the consistency between the graph pairs, while the feature augmentation network can preserve data semantics with the encoder perturbation.
Therefore, it is critical to combine data augmentation with feature augmentation in order to learn generalizable features.

\begin{figure}[!t]
	\centering
	\includegraphics[width=0.46\textwidth]{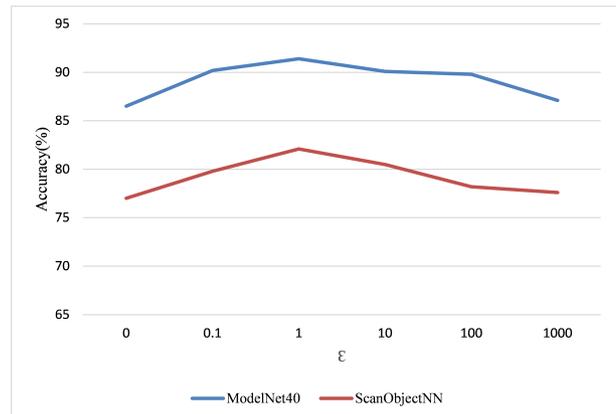}\\
	\caption{Effect of magnitude of the perturbation for the task of object classification on ModelNet40 and ScanObjectNN datasets in terms of accuracy.}
	\label{fig:perturbation}
\end{figure}

\textbf{Magnitude of the perturbation.} We then explore the magnitude of the perturbation which influences the learned representations.
As can be observed from Figure \ref{fig:perturbation}, if we set the magnitude of the perturbation as zero ($\epsilon$= 0), the performance is the lowest compared with other setting of perturbation on both  ModelNet40 and ScanObjectNN datasets.
Without perturbation, the AFSRL simply compares two
original samples as a negative pair and ignores the positive pair loss. 
Instead, appropriate perturbations encourage the model to learn representations
invariant to the perturbations through maximizing the consistency between a graph and the perturbed version.
It has come to our attention that increasing the amplitude of the perturbation while remaining within an adequate range can bring about a consistent performance improvement. 
On the other hand, overly significant perturbations will result in a decrease in performance since the semantics of the graph data will not be maintained.

\textbf{Transferability.}
Next, we aim to verify that our network has the ability to transfer the pre-trained models to other data domains using different pre-trained datasets. 
We pre-train the model on the existing largest natural dataset ScanNet \cite{dai2017scannet} and synthetic data ShapeNet \cite{chang2015shapenet}, and test their generalizability to different domains.
Table \ref{tab:Pre-train} reports the cross-domain experimental results, demonstrating
the successful transfer from models pre-trained on natural scenes to the synthetic shape domain.
The classification task achieves better performance when using ShapeNet as the pre-trained datasets compared with using ScanNet as pre-trained datasets for both ModelNet40 and ScanObjectNN datasets.
This is because the ShapeNet dataset provides point clouds with clean spatial structures and fewer noises, which is beneficial to the pre-trained model to learn effective representations.

\begin{table}[]
	\renewcommand\arraystretch{2.0}
	\caption{Linear evaluation for shape classification across different pre-train datasets.}
	\centering
	\begin{tabular}{l|c|c}
		\hline
		Dataset & Pre-train Dataset               &  Accuracy (\%) \\ \hline
		
		ModelNet40    &       ScanNet                                 &    90.7      \\
		ModelNet40   &            ShapeNet                            &  \textbf{91.5}      \\
		\hline
		
		ScanObjectNN   &ScanNet                 &  81.7       \\
		ScanObjectNN   &ShapeNet                 &    \textbf{82.3}     \\
		
		\hline
	\end{tabular}
	\label{tab:Pre-train}
\end{table}

\textbf{Generalizability.} We validate the practical use of AFSRL using GCN \cite{kipf2016semi}, GAT \cite{velivckovic2017graph}, and GRAPHSAGE \cite{hamilton2017inductive} as the encoders, respectively.
The flexibility of the proposed architecture allows us to seamlessly integrate it into different kind of GNN architectures.
Table \ref{tab:GNN} reports the classification results on ModelNet40 and ScanObjectNN datasets.
Note that the GCN outperforms the other two backbones as the encoder.

\begin{table}[]
	\renewcommand\arraystretch{2.0}
	\caption{Linear evaluation for shape classification with different GNN encoders.}
	\centering
	\begin{tabular}{l|c|c}
		\hline
		Model & ModelNet40            &  ScanObjectNN \\ \hline
		
		GRAPHSAGE \cite{hamilton2017inductive}    &       89.7                                 &    80.5      \\
		GAT \cite{velivckovic2017graph}  &            90.7                            &  81.1      \\

		GCN \cite{kipf2016semi}  &\textbf{91.5}                 &    \textbf{82.3}     \\
		
		\hline
	\end{tabular}
	\label{tab:GNN}
\end{table}

\section{Conclusion}
\label{sec:Conclusion}
In this paper, we propose a self-supervised representation learning framework on point clouds. 
Unlike prior methods that rely on heuristics for data augmentation, our method learns the representation by the combination of data augmentation and feature augmentation.
We validate our representation learning method on object classification and part segmentation tasks, and make extensive comparisons with the state-of-the-art approaches. 
The experimental results show that AFSRL can achieve superior performance compared to existing self-supervised methods. 
Although in this paper we only implement AFSRL on two tasks, AFSRL can also be applied to other problems with appropriate transformations on the data sets. We intend to investigate these issues in future works.


\section*{Acknowledgement}

The authors sincerely acknowledge the anonymous reviewers for their insights and comments to further improve the quality of the manuscript.
They also would like to thank the participants in the study for their valuable time.




%
%
%
%
%
%
%
%
\bibliographystyle{cas-model2-names}
%
\bibliography{cas-dc-template}
%
%

\end{document}